\documentclass{article}

\PassOptionsToPackage{numbers}{natbib}
\usepackage[final]{nips_2017}

\usepackage[utf8]{inputenc} % allow utf-8 input
\usepackage[T1]{fontenc}    % use 8-bit T1 fonts
\usepackage{hyperref}       % hyperlinks
\usepackage{url}            % simple URL typesetting
\usepackage{booktabs}       % professional-quality tables
\usepackage{amsfonts}       % blackboard math symbols
\usepackage{nicefrac}       % compact symbols for 1/2, etc.
\usepackage{microtype}      % microtypography

\usepackage{times}
\usepackage{graphicx}
\usepackage{subcaption} 
\usepackage{algorithm}
\usepackage{algorithmic}
\usepackage{amsmath,amsfonts,amssymb,amsthm}
\usepackage{multirow}
\usepackage{appendix}

\theoremstyle{plain}

\newtheorem{defn}{Definition}

\newtheorem{theorem}{Theorem}

\theoremstyle{definition}
\newtheorem{remark}{Remark}

\newtheorem*{rep@theorem}{\rep@title}
\newcommand{\newreptheorem}[2]{%
	\newenvironment{rep#1}[1]{%
		\def\rep@title{#2 \ref{##1}}%
		\begin{rep@theorem}}%
		{\end{rep@theorem}}}
\newreptheorem{theorem}{Theorem}
\newreptheorem{lemma}{Lemma}
\newreptheorem{corollary}{Corollary}
\newreptheorem{prop}{Proposition}

\newcommand{\R}{\mathbb{R}}

\newcommand{\cX}{\mathcal{X}}
\newcommand{\cC}{\mathcal{C}}
\newcommand{\cS}{\mathcal{S}}

\newcommand{\iIF}[1]{\STATE\algorithmicif\ #1\ \algorithmicthen}
\newcommand{\iELSIF}[1]{\STATE\algorithmicelsif\ #1\ \algorithmicthen}
\newcommand{\iELSE}[1]{\STATE\algorithmicelse\ #1}
\newcommand{\iENDIF}{\unskip\STATE\algorithmicend\ \algorithmicif}

\title{Relaxed Oracles for Semi-Supervised Clustering}

\author{
  Taewan~Kim \\
  The University of Texas at Austin\\
  \texttt{twankim@utexas.edu}
  \And
  Joydeep~Ghosh\\
  The University of Texas at Austin\\
  \texttt{jghosh@utexas.edu}
}

\begin{document}
% \nipsfinalcopy is no longer used

\maketitle

%%%%%%%%%%%%%%%%%%%% Abstract %%%%%%%%%%%%%%%%%%%%
\begin{abstract}
	 Pairwise ``same-cluster'' queries are one of the most widely used forms of supervision in semi-supervised clustering. However, it is impractical to ask human oracles to answer every query correctly. In this paper, we study the influence of allowing ``not-sure'' answers from a weak oracle and propose an effective algorithm to handle such uncertainties in query responses. Two realistic weak oracle models are considered where ambiguity in answering depends on the distance between two points. We show that a small query complexity is adequate for effective clustering with high probability by providing better pairs to the weak oracle. Experimental results on synthetic and real data show the effectiveness of our approach in overcoming supervision uncertainties and yielding high quality clusters. \footnote{This paper focuses on the distance-based weak oracle models with additional experimental results. Proofs for theoretical results are available in the extended version. \cite{kim2017semi}}
\end{abstract}

%%%%%%%%%%%%%%%%%%%% Introduction %%%%%%%%%%%%%%%%%%%%
\section{Introduction}\label{sec:intro}
Clustering is one of the most popular procedures for extracting meaningful insights from unlabeled data. However, clustering is also very challenging for a wide variety of reasons \cite{jain1999data}. Finding the optimal solution of even the simple $k$-means objective is known to be NP-hard \cite{davidson2005clustering,mahajan2009planar,vattani2009hardness,reyzin2012data}. Second, the quality of a clustering algorithm is difficult to evaluate without context. Semi-supervised clustering is one way to overcome these problems by providing a small amount of additional knowledge related to the task \cite{basu2002semi,cohn2003semi,basu2004active,basu2004probabilistic,balcan2008clustering,kulis2009semi,mazumdar2016clustering,ashtiani2016clustering,mazumdar2017query,ailon2017approximate}.

The semi-supervised active clustering (SSAC) framework proposed by \citet{ashtiani2016clustering} combines both margin property and pairwise constraints in the active query setting. A domain expert can help clustering by answering same-cluster queries, which ask whether two samples belong to the same cluster or not. By using an algorithm with two phases, it was shown that the oracle's clustering can be recovered in polynomial time with high probability. However, their formulation of the same-cluster query has only two choices of answers, \textit{yes} or \textit{no}. This might be impractical as a domain expert can also encounter ambiguous situations which are difficult to respond to in a short time.

Our work is motivated by the following question: ``Is it possible to perform a clustering task efficiently even with a non-ideal domain expert?''. We answer this question by formulating practical weak oracle models and allowing \textit{not-sure} answers to query responses. Our model assumptions considers two reasonable scenarios that may lead to ambiguity in answering a same-cluster query: (i) distance between two points from different clusters is too small, and (ii) distance between two points within the same cluster is too large. We prove that our improved SSAC algorithm can work well under uncertainties if there exists at least one cluster element close enough to the center.

Experimental results on both synthetic and real data show the effective performance of our approach. In particular, our algorithm successfully deals with uncertainties compared to the previous SSAC algorithm by relaxing an oracle's role and providing better pairs for annotation in an active semi-supervision framework.

%%%%%%%%%%%%%%%%%%%%% Section: Background problem Formulation %%%%%%%%%%%%%%%%%%%%%

\section{Problem Setting}\label{sec:problemsetup}

For the purpose of theoretical analysis, the domain of data is assumed to be the Euclidean space $\R^m$, and each center of a clustering $\cC$ is defined as a mean of elements in the corresponding cluster, i.e. $\mu_i = \frac{1}{|C_i|}\sum_{x\in C_i}x, \forall i\in[k]$. Then, an optimal solution of the $k$-means clustering is a center-based clustering.\footnote{In fact, this will hold for all Bregman divergences \cite{banerjee2005clustering}.} Also, a $\gamma$-margin property ensures the existence of an optimal clustering.

\begin{defn}[Center-based clustering]\label{def:cb_clustering}
	A clustering $\cC = \{C_1,\cdots,C_k\}$ is a center-based clustering of $\cX \subset \R^m$ with $k$ clusters, if there exists a set of centers $\mu = \{\mu_1,\cdots,\mu_k\}\subset \R^m$ satisfying the following condition with a distance metric $d(x,y)$:\\
	\centerline{$x\in C_i \Leftrightarrow i = \arg\min_{j} d(x,\mu_j),~~\forall x\in\cX$  and $i \in [k]$}
\end{defn}
\begin{defn}[$\gamma$-margin property - Clusterability]\label{def:gamma}
	Let $\cC$ be a center-based clustering of $\cX$ with clusters $\cC =\{C_1,\cdots,C_k\}$ and corresponding centers $\{\mu_1, \cdots,\mu_k\}$. $\cC$ satisfies the $\gamma$-margin property if the following condition is true:\\
	\centerline{$\gamma d(x,\mu_i)<d(y,\mu_i),~~\forall i\in[k], \forall x\in C_i, \forall y\in \cX\setminus C_i$}
\end{defn}

\paragraph{Problem Formulation}\label{subsec:problem} We apply the SSAC algorithm on data $\cX$, which is supported by a weak oracle that receives weak same-cluster queries. The true clustering $\cC$ satisfies the $\gamma$-margin property.
\begin{defn}[Weak Same-cluster Query]\label{def:weak_query}
	A weak same-cluster query asks whether two data points $x_1,x_2 \in \cX$ belong to the same cluster and receives one of three responses from an oracle.
	\begin{align*}
	Q(x_1,x_2) = \begin{cases}
	1&\text{if } x_1, x_2\text{ are in the same cluster}\\
	0&\text{if not-sure}\\
	-1&\text{if } x_1, x_2\text{ are in different clusters}
	\end{cases}
	\end{align*}
\end{defn}

\begin{defn}[Weak Pairwise Cluster-assignment Query]\label{def:weak_assign_query}
	A weak pairwise cluster-assignment query identifies the cluster index of a given data point $x$ by asking $k$ weak same-cluster queries $Q(x,y_i)$, where $y_i \in C_{\pi(i)},~i\in[k]$. One of $k+1$ responses is inferred from an oracle with $\cC=\{C_1,\cdots,C_k\}$. $\pi(\cdot)$ is a permutation defined on $[k]$ which is determined during the assignment process accordingly.
	\begin{align*}
	Q(x) = \begin{cases}
	t&\text{if } x\in C_{\pi(t)}, t\in[k]\\
	0&\text{if not-sure}\\
	\end{cases}
	\end{align*}
\end{defn}
In our framework, the cluster-assignment process uses $k$ weak same-cluster queries and therefore only depends on pairwise information provided by weak oracles. And we denote the radius of a cluster as $r(C_i)\triangleq \max_{x \in C_i}d(x,\mu_i)$ throughout the paper.

\begin{algorithm} [ht]
	\caption{SSAC for Weak Oracles}
	\label{alg:weak_SSAC}
	\begin{algorithmic} [1] % enter the algorithmic environment
		\REQUIRE Dataset $\cX$, an oracle for weak query $Q$, target number of clusters $k$, sampling numbers $(\eta,\beta)$, and a parameter $\delta\in(0,1)$.
		\STATE $\cC=\{\},~~\cS_1 = \cX,~~r=\lceil k\eta\rceil$
		\FOR{$i=1$ to $k$}
		\STATE \vspace{.2em}\textbf{- Phase 1:}
		\begin{ALC@g}
		\STATE $Z\sim \text{Uniform}(\cS_i,r)$ \hspace{2em}// Draw $r$ samples from $S_i$
		\FOR{$1\leq t \leq k$} 
		\STATE $Z_t=\{x\in Z:Q(x)=t\}$ \hspace{2em}// Pairwise cluster-assignment query
		\ENDFOR
		\STATE $p=\arg\max_t |Z_t|$, $\mu_p' \triangleq \frac{1}{|Z_p|}\sum_{x\in Z_p} x$
		\end{ALC@g}
		\STATE \vspace{.2em}\textbf{- Phase 2:}
		\begin{ALC@g}
		\STATE $\hat{\cS_i} = \text{sorted}(\cS_i)$ \hspace{2em}// Increasing order of $d(x,\mu_p'),~x\in\cS_i$
		\STATE $r_i' =$ BinarySearch($\hat{\cS_i},Z_p,\mu_p',\beta$)\hspace{2em}// Same-cluster query
		\STATE $C_p'=\{x\in \cS_i: d(x,\mu_p') < r_i' \},~~\cS_{i+1} = \cS_i \setminus C_p',~~\cC=\cC \cup \{C_p'\}$
		\end{ALC@g}
		\ENDFOR
		\ENSURE A clustering $\cC$ of the set $\cX$
	\end{algorithmic}
\end{algorithm}

%%%%%%%%%%%%%%%%%%%%% Section: Distance-based weak oracle %%%%%%%%%%%%%%%%%%%%%%%

\section{SSAC with Distance-Weak Oracles}\label{sec:weak_oracle_dist}
It is reasonable to expect the accuracy of feedback from domain experts to depend on the inherent ambiguities of the given pairs of samples. The cause of ``not-sure'' answer for the same-cluster query can be investigated based on the distance between the elements in a feature space. Two reasons for having indefinite answers are considered in this work: (i) points from different clusters are too close, and (ii) points within the same cluster are too far. The first situation happens a lot in the real world. For instance, distinguishing wolves from dogs is not an easy task if a Siberian Husky is considered. The second case is also reasonable, because it might be difficult to compare characteristics of two points within the same cluster if they have quite dissimilar features.

\begin{algorithm} [ht]
	\caption{Unified-Weak BinarySearch}
	\label{alg:unified_Binary}
	\begin{algorithmic} [1] % enter the algorithmic environment
		\REQUIRE Sorted dataset $\hat{\cS_i}=\{x_1,\cdots,x_{|\hat{\cS_i}|}\}$ in increasing order of $d(x_j,\mu_p')$, an oracle for weak query $Q$, target cluster $p$, set of assignment-known points $Z_p$, empirical mean $\mu_p'$, and a sampling number $\beta \leq |Z_p|$.
		\STATE \textbf{- Search}($x_j\in\hat{\cS_i}$)\textbf{:}
		\begin{ALC@g}
		\STATE Select the point $x_1$ and use it for same-cluster queries
		\iIF{$Q(x_1,x_j)=1$} Set left bound index as $j+1$
		\iELSIF{$Q(x_1,x_j)=-1$} Set right bound index as $j-1$
		\iELSE
		\begin{ALC@g}
		\STATE Sample $\beta-1$ points from $Z_p$. $B\subseteq Z_p,~|B|=\beta-1$\\
		\STATE Weak same-cluster query $Q(x_j,y)$, for all $y\in B$
		\iIF{$x_j$ is in cluster $C_p$} Set left bound index as $j+1$
		\iELSE Set right bound index as $j-1$
		\iENDIF
		\end{ALC@g}
		\iENDIF
		\end{ALC@g}
		\STATE \vspace{.2em}\textbf{- Stop:}\hspace{.5em}Found the smallest index $j^*$ such that $x_{j^*}$ is not in $C_p$
		\ENSURE $r_i'=d(x_{j^*},\mu_p')$		
	\end{algorithmic}
\end{algorithm}
\begin{remark}
	Algorithm \ref{alg:unified_Binary} can also handle oracles with a random behavior. $\beta=1$ is sufficient for distance-weak oracles.
\end{remark}

\begin{figure}[ht]
	\begin{center}
		\centerline{\includegraphics[width=.7\linewidth]{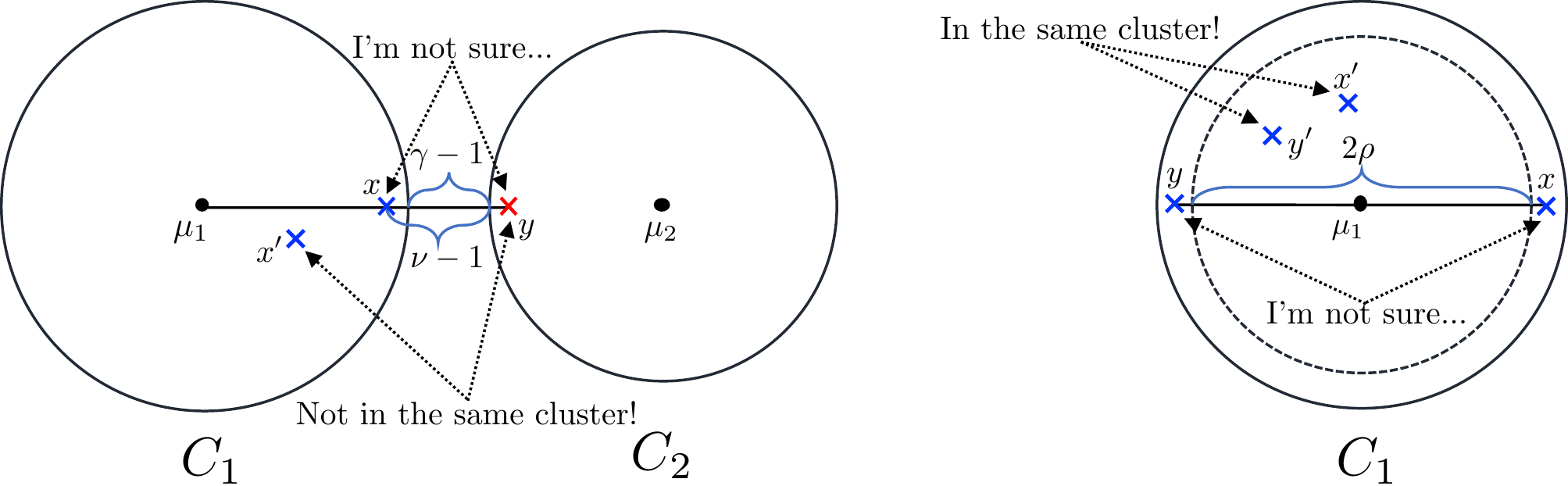}}
		\caption{Visual representation of two \textit{not-sure} cases for the local distance-weak oracle. (\textit{Left}) Two points from the different clusters are too close. (\textit{Right}) Two points from the same clusters are too far.}
		\label{fig:local_dist_weak}
	\end{center}
	\vskip -0.1in
\end{figure}

\paragraph{Local Distance-Weak Oracle}\label{subsec:dist_weak_local} We define the first weak-oracle model sensitive to distance, a local distance-weak oracle, in a formal way to include two vague situations described before. These confusing cases for local distance-weak oracle are visually depicted in Figure \ref{fig:local_dist_weak} for better explanation.

\begin{defn}[Local Distance-Weak Oracle]\label{def:weak_oracle_dist_local}
	An oracle having a clustering $\cC=\{C_1,\cdots,C_k\}$ for data $\cX$ is said to be $(\nu,\rho)$ local distance-weak with parameters $\nu\geq 1$ and $\rho\in(0,1]$, if $Q(x,y)=0$ for any given two points $x,y \in \cX$ satisfying one of the following conditions:
	\vspace{-.4em}
	\begin{align*}
	&\text{(a)}~d(x,y)<(\nu-1)\min\{d(x,\mu_i),d(y,\mu_j)\},\text{ where } x\in C_i, y\in C_j, i\neq j\\
	&\text{(b)}~d(x,y)>2\rho r(C_i),\text{ where }x,y\in C_i
	\end{align*}
\end{defn}
One way to overcome the uncertainty is to provide at least one good point in a query, i.e. \textit{better pairs}. If one of the points $x$ and $y$ for the query $Q(x,y)$ is close enough to the center of a cluster, a local distance-weak oracle does not get confused in answering. This situation is realistic because one representative data sample of a cluster might be a good baseline when comparing to other elements. Theorem \ref{thm:optimal_cover_local} is founded on this intuition, and we show that our modified version of SSAC will succeed if at least one representative sample per cluster is suitable for the weak oracle. 

\begin{theorem}\label{thm:optimal_cover_local}
	If a cluster $C_i$ contains at least one point $x^*\in C_i$ satisfying $d(x^*,\mu_i)<c_{local}\cdot r(C_i)$ for all $i\in[k]$, then combination of Algorithm \ref{alg:weak_SSAC} and \ref{alg:unified_Binary} outputs the oracle's clustering $\cC$ with probability at least $1-\delta$ by asking weak same-cluster queries to a $(\nu,\rho)$ local distance-weak oracle. $\left( c_{local}=\min\{2\rho-1,\gamma-\nu+1\}-2\epsilon\right.$, where $\left.\epsilon\leq \frac{\gamma-1}{2}\right)$
\end{theorem}
{\it Sketch of Proof.} We first show the effect of a point close to the center on weak queries. Then the possibility of having a close empirical mean is provided by defining \textit{good sets} and calculating data-driven probability of failure from it. Last, an assignment-known point is identified to remove the uncertainty of same-cluster queries used in the binary search step.

\paragraph{Global Distance-Weak Oracle}\label{subsec:dist_weak_global} A global distance-weak oracle fails to answer depending on the distance of each point to its respective cluster center. In this case, both elements $x$ and $y$ should be in the covered range of an oracle if they don't belong to the same cluster.

\begin{figure}[ht]
	\begin{center}
		\centerline{\includegraphics[width=.55\linewidth]{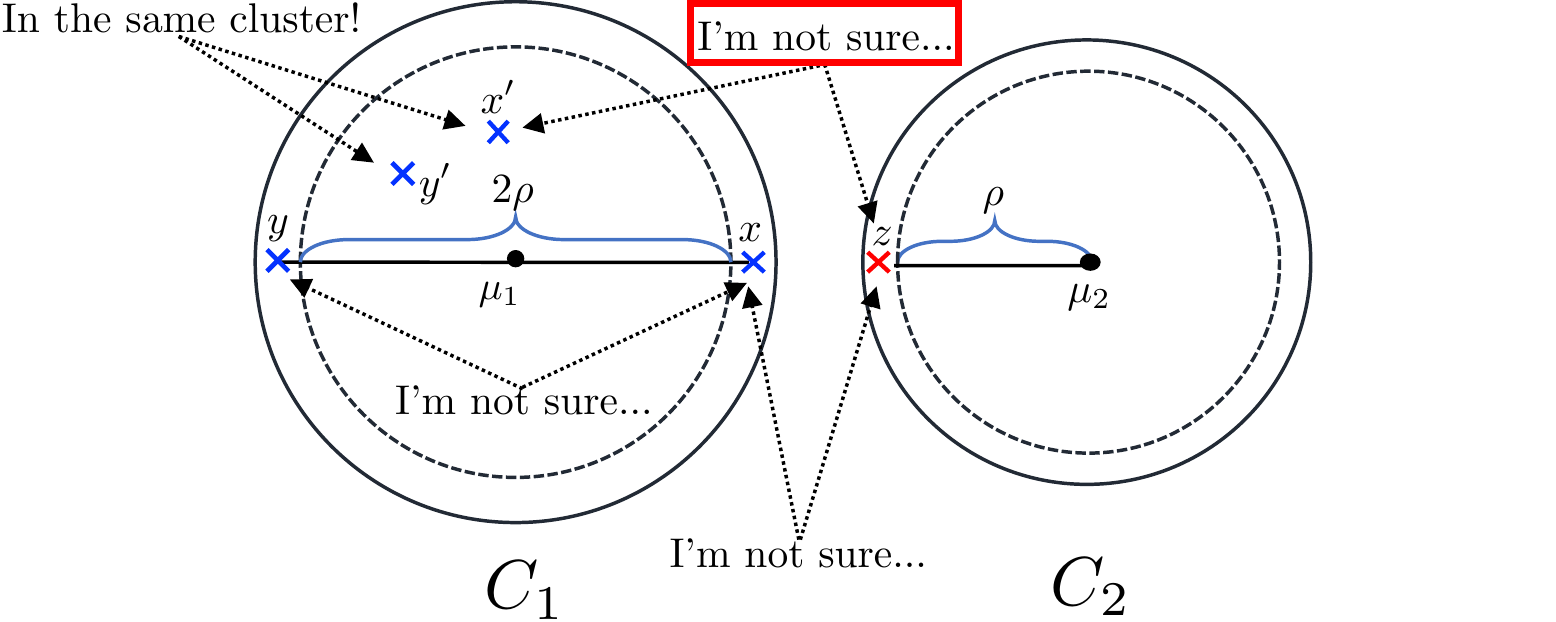}}
		\caption{Visual representation of two \textit{not-sure} cases for the global distance-weak oracle. The red box indicates the difference with the local distance-weak oracle.}
		\label{fig:global_dist_weak}
	\end{center}
	\vskip -0.1in
\end{figure}
\begin{defn}[Global Distance-Weak Oracle]\label{def:weak_oracle_dist_global}
	An oracle having a clustering $\cC=\{C_1,\cdots,C_k\}$ for data $\cX$ is said to be $\rho$ global distance-weak with parameter $\rho\in(0,1]$, if $Q(x,y)=0$ for any given two points $x,y \in \cX$ satisfying one of the following conditions:
	\vspace{-.4em}
	\begin{align*}
	&\text{(a)}~d(x,\mu_i)>\rho r(C_i)\text{ or }d(y,\mu_j)>\rho r(C_j),\text{ where }x\in C_i, y\in C_j, i\neq j\\
	&\text{(b)}~d(x,y)>2\rho r(C_i),\text{ where }x,y\in C_i
	\end{align*}
\end{defn}
The problem of a global distance-weak oracle compared to the local distance-weak model is the increased ambiguity in distinguishing elements from different clusters. Nevertheless, once we get a good estimate of the center, better pairs with one good point can be still found to support the oracle in answering same-cluster queries.

\begin{theorem}\label{thm:optimal_cover_global}
	If a cluster $C_i$ contains at least one point $x^*\in C_i$ satisfying $d(x^*,\mu_i)<c_{global}\cdot r(C_i)$ for all $i\in[k]$, then combination of Algorithm \ref{alg:weak_SSAC} and \ref{alg:unified_Binary} outputs the oracle's clustering $\cC$ with probability at least $1-\delta$, by asking weak same-cluster queries to a $\rho$ global distance-weak oracle. $\left(c_{global}=2\rho-1-2\epsilon\right.$, where $\left.\epsilon\leq \frac{\gamma-1}{2}\right)$
\end{theorem}

%%%%%%%%%%%%%%%%%%%%% Experimental Results %%%%%%%%%%%%%%%%%%%%%

\section{Experimental Results}\label{sec:exp_results}

\paragraph{Synthetic Data}\label{subsec:exp_syn_data} Points of each cluster are generated from isotropic Gaussian distribution. We assume that there exists a ground truth oracle's clustering, and the goal is to recover it where labels are partially provided via weak same-cluster queries. For visual representation, 2-dimensional data points are considered, and other parameters are set to $n=600$ (number of points), $k=3$ (number of clusters), and $\sigma_{std}=2.0$. Data points satisfy $\gamma$-margin property with condition $\gamma_{\min}\leq \gamma \leq \gamma_{\max}$. To focus on scenarios with narrow margins, $\gamma_{\min}=1.0$ and $\gamma_{\max}=1.1$ are chosen.

\paragraph{MNIST}\label{subsec:exp_mnist} $\gamma$-margin property is difficult to evaluate and satisfy in real world data as a \textit{good} representation or an embedding space is not given. Therefore, we assumed that the oracle has a 2-dimensional embedding space equivalent to the one generated by t-Distributed Stochastic Neighbor Embedding (t-SNE) algorithm \cite{van2014accelerating}. We used digits 0, 6, and 8 in the subset of MNIST dataset for similarity.\footnote{Sample MNIST (2500 points) is from the t-SNE code. \url{https://lvdmaaten.github.io/tsne/}}

\paragraph{Evaluation}\label{subsec:exp_eval}
Each round of the evaluation is composed of experiments with different parameter settings on $(\eta,c_{dist})$. Parameters for the distance-weak oracles, $\rho$ and $\nu$, are controlled by $c_{dist}$ in the experiments: $\rho=c_{dist}$ and $\nu=\max(1,\gamma)+2\cdot(1-c_{dist})$. $\beta$ is fixed as $1$ since we are only considering distance-weak oracles. $\eta$ and $c_{dist}$ are varied in each round, and the task is repeated $5000$ (MNIST) and $10000$ (Synthetic) times. Two evaluation metrics are considered: $Accuracy$ is the ratio of correctly recovered data points averaged over $n$ points, and $\# Failure$ is the total number of failures occurred at cluster-assignments. The best permutation for the cluster labels is investigated based on the distances between estimated centers and true centers for the evaluation. To compare the performance of our improved SSAC, the original one \cite{ashtiani2016clustering} receives random answers, $Q(x,y)=\pm 1$ with probability 0.5, whenever an oracle encounters the case of not-sure. Also, pairs used in the binary search steps are randomly selected from the cluster-known points.

\paragraph{Results}\label{subsec:exp_results}
\begin{figure}[t]
	\centering
	\begin{subfigure}{.24\linewidth}
		\centering
		\includegraphics[width=\linewidth]{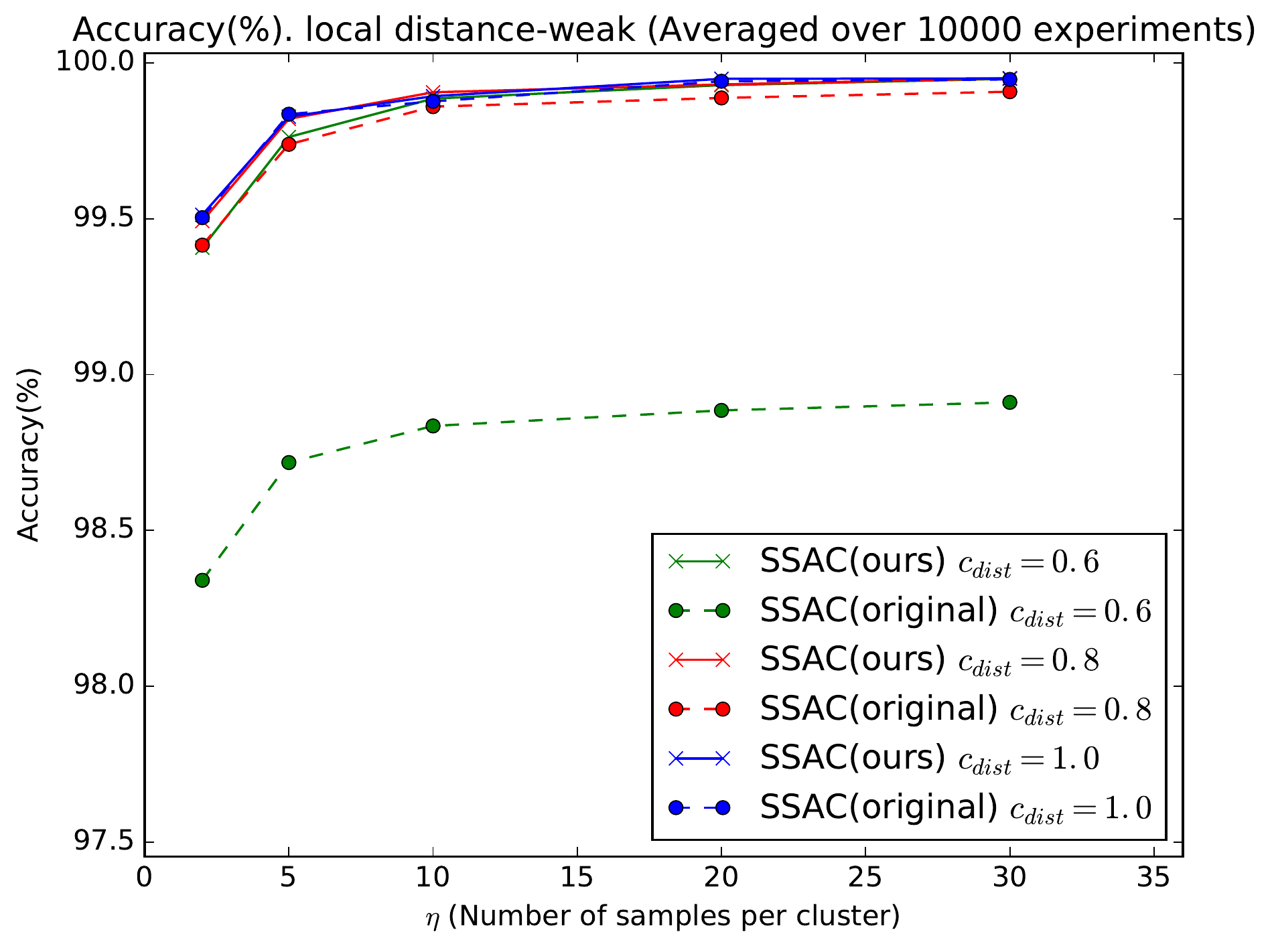}
		\caption{$Accuracy~(\%)$}\label{fig:res_syn_local_acc}
	\end{subfigure}
	\begin{subfigure}{.24\linewidth}
		\centering
		\includegraphics[width=\linewidth]{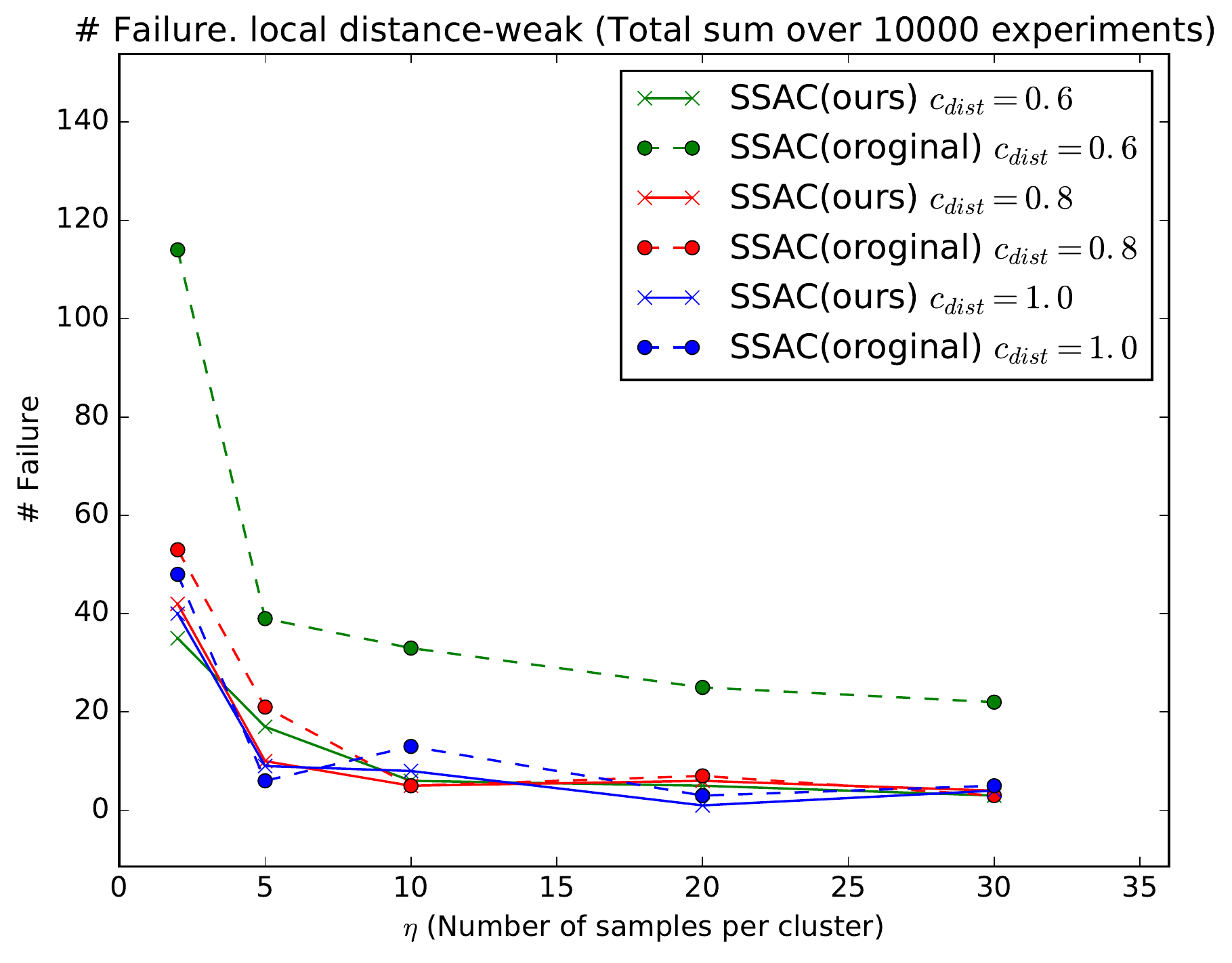}
		\caption{$\#~Failure$}\label{fig:res_syn_local_fail}
	\end{subfigure}
	\hspace{.5em}
	\begin{subfigure}{.24\linewidth}
		\centering
		\includegraphics[width=\linewidth]{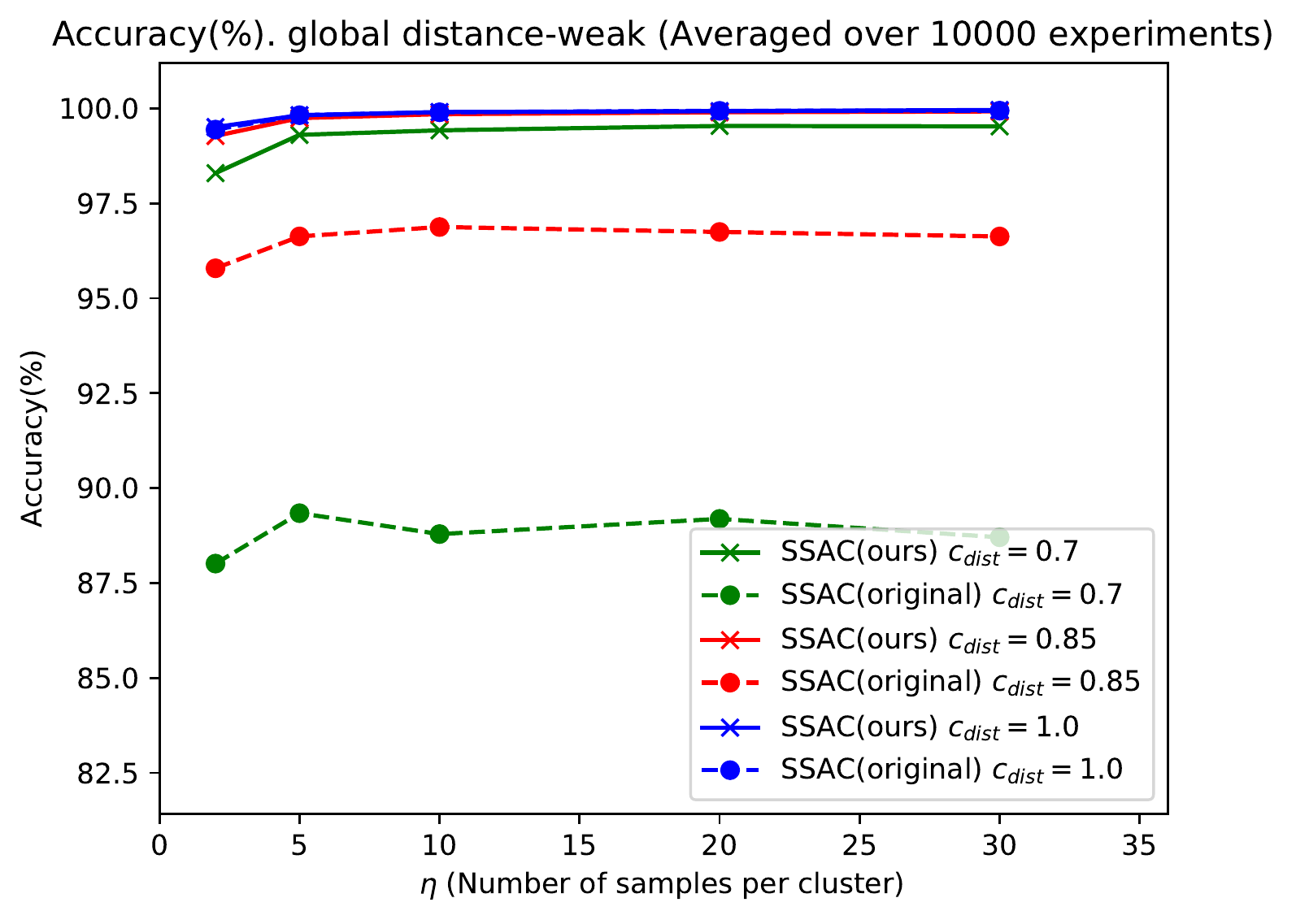}
		\caption{$Accuracy~(\%)$}\label{fig:res_syn_global_acc}
	\end{subfigure}
	\begin{subfigure}{.24\linewidth}
		\centering
		\includegraphics[width=\linewidth]{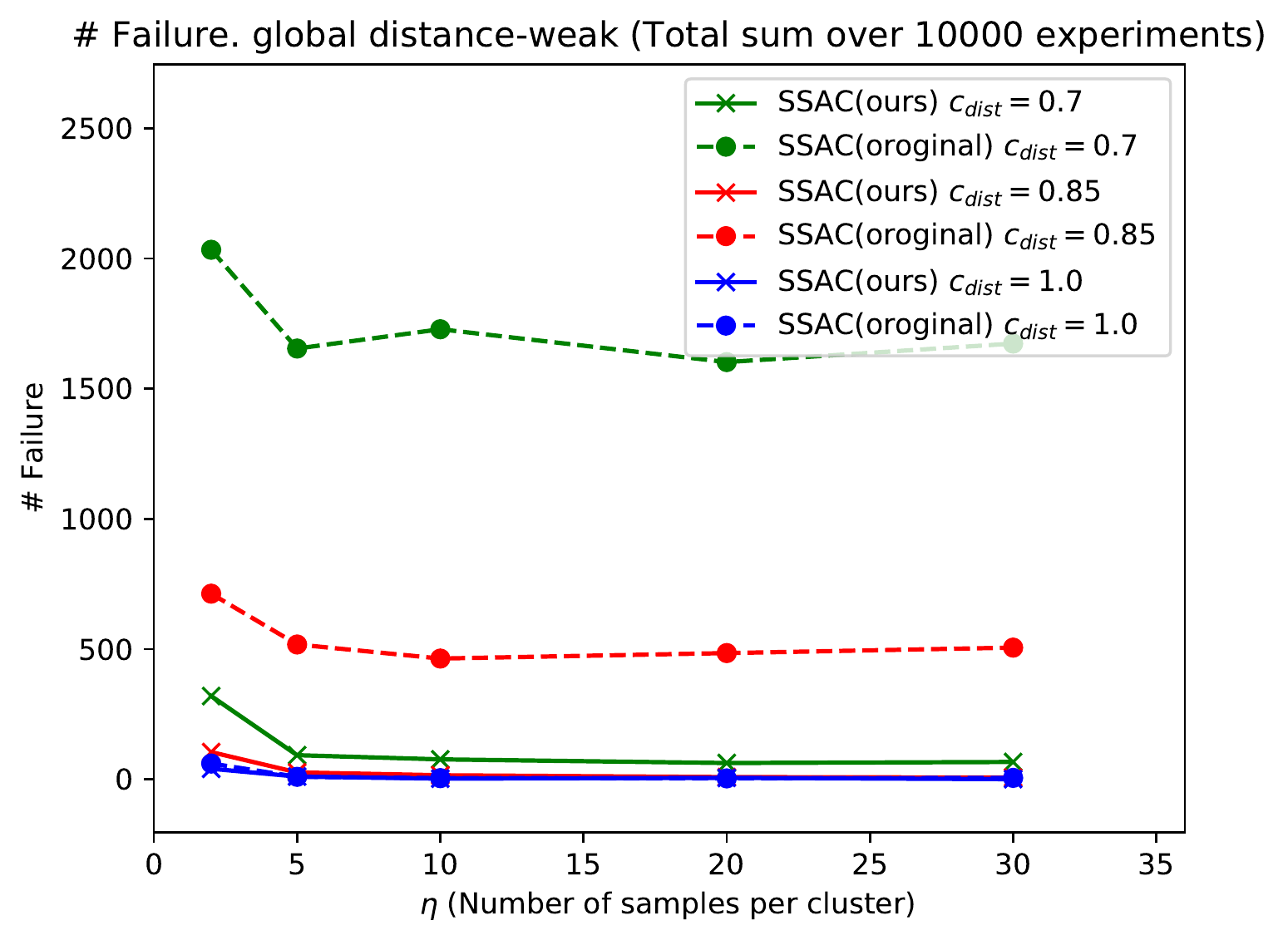}
		\caption{$\#~Failure$}\label{fig:res_syn_global_fail}
	\end{subfigure}
	\caption{Synthetic data. (\subref{fig:res_syn_local_acc}),(\subref{fig:res_syn_local_fail}): Local distance-weak oracle, $c_{dist}\in \{0.6,0.8,1.0\}$. (\subref{fig:res_syn_global_acc}),(\subref{fig:res_syn_global_fail}): Global distance-weak oracle, $c_{dist}\in \{0.7,0.85,1.0\}$. \textit{x-axis}: $\eta\in\{2,5,10,20,30\}$ (Number of samples)}
	\label{fig:res_syn}
	\vskip -0.1in
\end{figure}
\begin{figure}[t]
	\centering
	\begin{subfigure}{.24\linewidth}
		\centering
		\includegraphics[width=\linewidth]{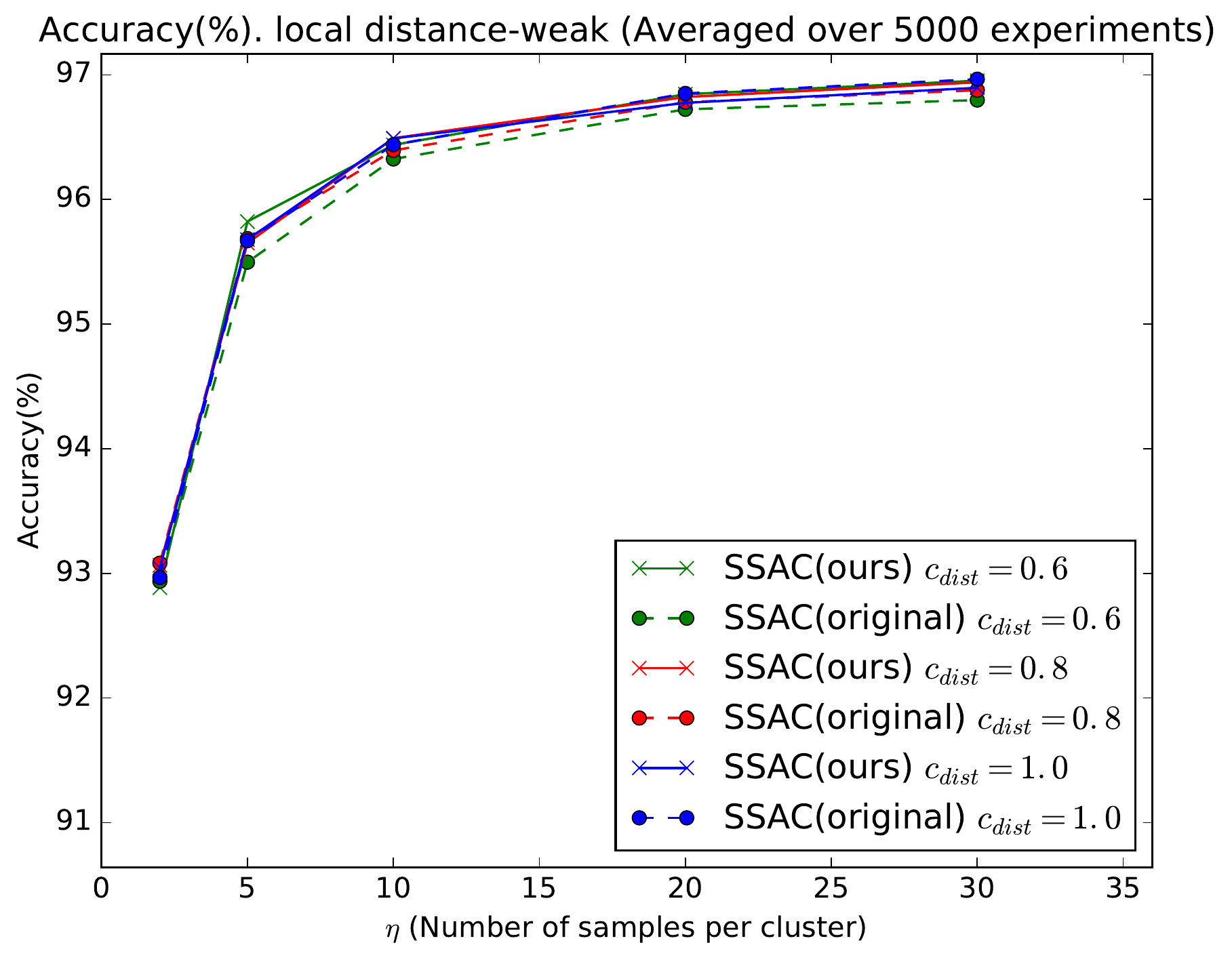}
		\caption{$Accuracy~(\%)$}\label{fig:res_mnist_local_acc}
	\end{subfigure}
	\begin{subfigure}{.24\linewidth}
		\centering
		\includegraphics[width=\linewidth]{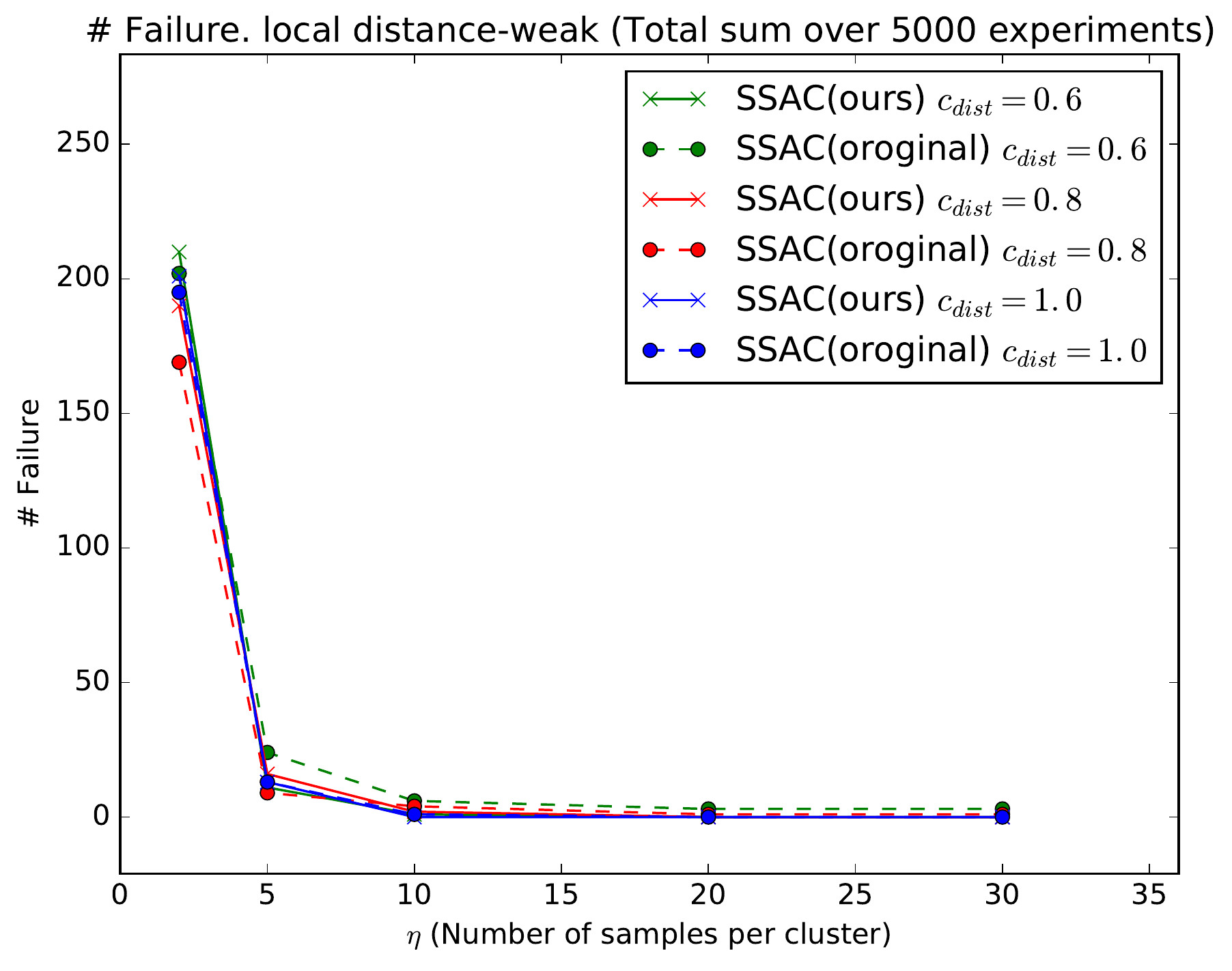}
		\caption{$\#~Failure$}\label{fig:res_mnist_local_fail}
	\end{subfigure}
	\hspace{.5em}
	\begin{subfigure}{.24\linewidth}
		\centering
		\includegraphics[width=\linewidth]{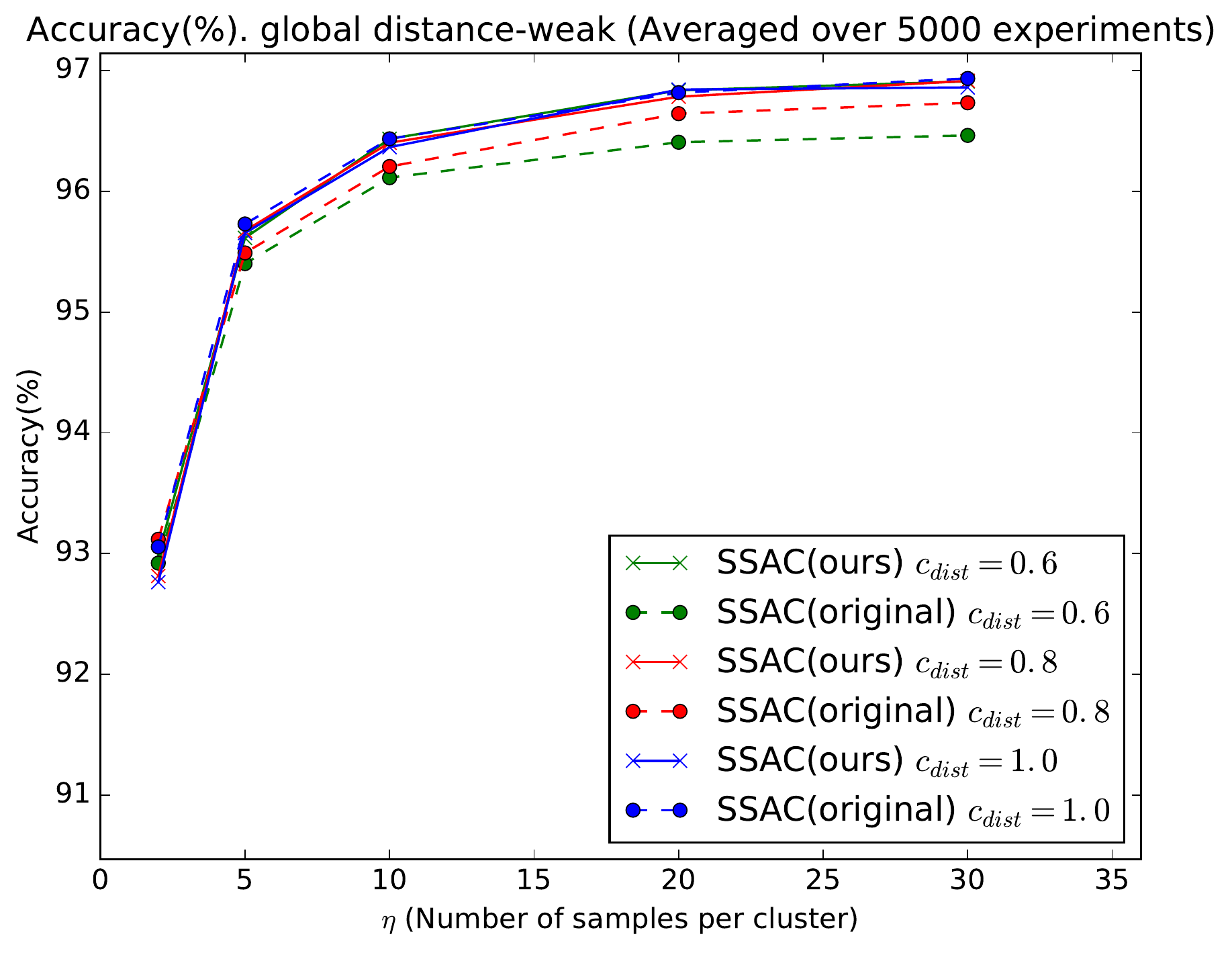}
		\caption{$Accuracy~(\%)$}\label{fig:res_mnist_global_acc}
	\end{subfigure}
	\begin{subfigure}{.24\linewidth}
		\centering
		\includegraphics[width=\linewidth]{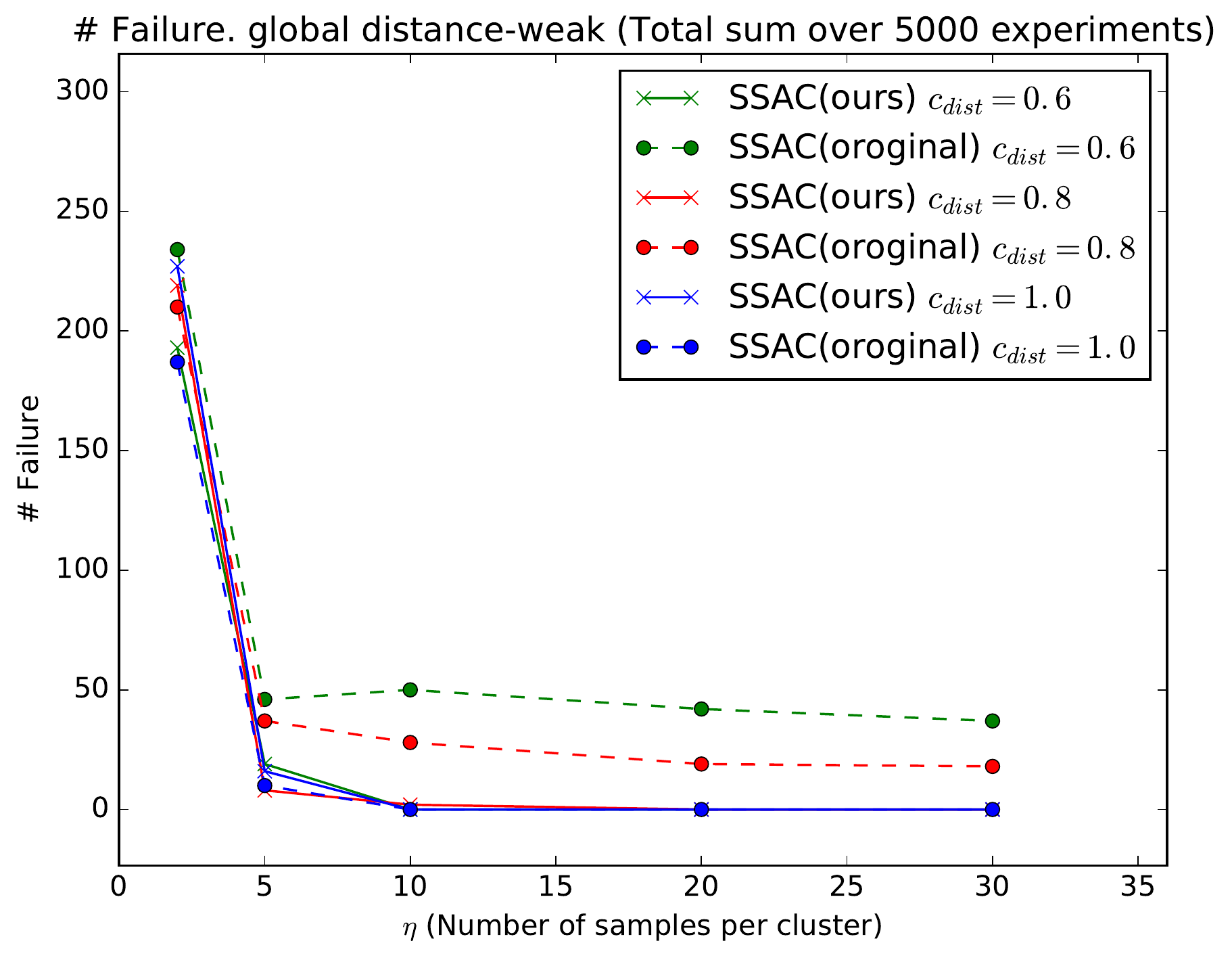}
		\caption{$\#~Failure$}\label{fig:res_mnist_global_fail}
	\end{subfigure}
	\caption{MNIST. (\subref{fig:res_mnist_local_acc}),(\subref{fig:res_mnist_local_fail}): Local distance-weak oracle, $c_{dist}\in \{0.6,0.8,1.0\}$. (\subref{fig:res_mnist_global_acc}),(\subref{fig:res_mnist_global_fail}): Global distance-weak oracle, $c_{dist}\in \{0.6,0.8,1.0\}$. \textit{x-axis}: $\eta\in\{2,5,10,20,30\}$ (Number of samples)}
	\label{fig:res_mnist}
	\vskip -0.1in
\end{figure}
An accuracy improves as $\eta$ increases, and this shows the importance of enough number of samples to succeed in clustering with weak oracles. In fact, even small number of samples are sufficient in practice. Failures of the SSAC algorithm can happen as it is a probabilistic algorithm. When $\eta$ is really small, the possibility of failure increases as we have only few chances to ask cluster-assignment queries. For example, if $\eta=2$, only $r=\lceil k\eta \rceil=6$ points are sampled. Then, if all 6 cluster-assignment queries fail, Phase 1 fails which leads to the recovery of less than $k$ clusters. However, such situations rarely occur if $\eta$ is large enough.

Results in Figure \ref{fig:res_syn} and \ref{fig:res_mnist} show that our improved algorithm (solid lines) outperforms the vanilla SSAC (dashed lines) by allowing not-sure query responses to relax oracles. Especially, results on synthetic data clearly prove the effectiveness of providing better pairs to weak oracles in binary search steps. Our algorithm is robust against the different level of distance weakness. Also, empirical results on MNIST further supports the practicality of our algorithm and weak models.\footnote{The source code is available online. \url{https://github.com/twankim/weaksemi}}

%%%%%%%%%%%%%%%%%%%%% Discussion/Conclusion %%%%%%%%%%%%%%%%%%%%%

\section{Conclusion and Future Work}\label{sec:conclusion}
This paper presents an approach for utilizing weak oracles in clustering. Specifically, we suggest two realistic types of domain experts who can provide an answer ``not-sure'' for the same-cluster query. For each model, probabilistic guarantee on discovering the oracle's clustering is provided based on our improved algorithm. In particular, a single element close enough to the cluster center mitigates ambiguous supervision by providing better pairs to an oracle. One interesting future direction is to accommodate embedding learning methods for the real-world clustering tasks.

%%%%%%%%%%%%%%%%%%%%% References %%%%%%%%%%%%%%%%%%%%%
\newpage
\nocite{tropp2012user,awasthi2012center,balcan2016clustering,ashtiani2015representation,ashtiani2015dimension}
\bibliographystyle{plainnat}
{\small\bibliography{kim2017distweak}}

\end{document}